\documentclass[sigconf]{acmart}

\AtBeginDocument{%
  \providecommand\BibTeX{{%
    \normalfont B\kern-0.5em{\scshape i\kern-0.25em b}\kern-0.8em\TeX}}}

\acmYear{2023}

\setcopyright{none}

\acmConference[HRI 2023 Workshop]{2023 ACM/IEEE International Conference on Human-Robot Interaction}{March 13, 2023}{Stockholm, Sweden}

\acmBooktitle{2023 ACM/IEEE International Conference on Human-Robot Interaction (HRI) Workshop on Advancing HRI Research and Benchmarking Through Open-Source Ecosystems, March 13, 2023, Stockholm, Sweden}

\acmDOI{}

\acmISBN{}

\begin{document}

\title{Preserving HRI Capabilities: Physical, Remote and Simulated Modalities in the SciRoc 2021 Competition}


\author{Vincenzo Suriani}
\email{{surname}@diag.uniroma1.it}
\orcid{0000-0003-1199-8358}
\author{Daniele Nardi}
\orcid{0000-0001-6606-200X}


\affiliation{
  \institution{Department of Computer, Control, and Management Engineering, Sapienza University of Rome}
  \streetaddress{Via Ariosto 25}
  \city{Rome}
  \country{Italy}
  \postcode{00185}
}



\renewcommand{\shortauthors}{Suriani et al.}

\begin{abstract}
  In the last years, robots are moving out of research laboratories to enter everyday life. Competitions aiming at benchmarking the capabilities of a robot in everyday scenarios are useful to make a step forward in this path. In fact, they foster the development of robust architectures capable of solving issues that might occur during the human-robot coexistence in human-shaped scenarios. One of those competitions is \emph{SciRoc} that, in its second edition, proposed new benchmarking environments. In particular, Episode 1 of \emph{SciRoc 2} proposed three different modalities of participation while preserving the Human-Robot Interaction (HRI), being a fundamental benchmarking functionality.  
  The \textit{Coffee Shop} environment, used to challenge the participating teams, represented an excellent testbed enabling for the benchmarking of different robotics functionalities, but also an exceptional opportunity for proposing novel solutions to guarantee real human-robot interaction procedures despite the Covid-19 pandemic restrictions. The developed software is publicly released.
\end{abstract}





\maketitle


\section{Introduction}
\begin{figure}
    \centering
    \includegraphics[width=1.0\columnwidth]{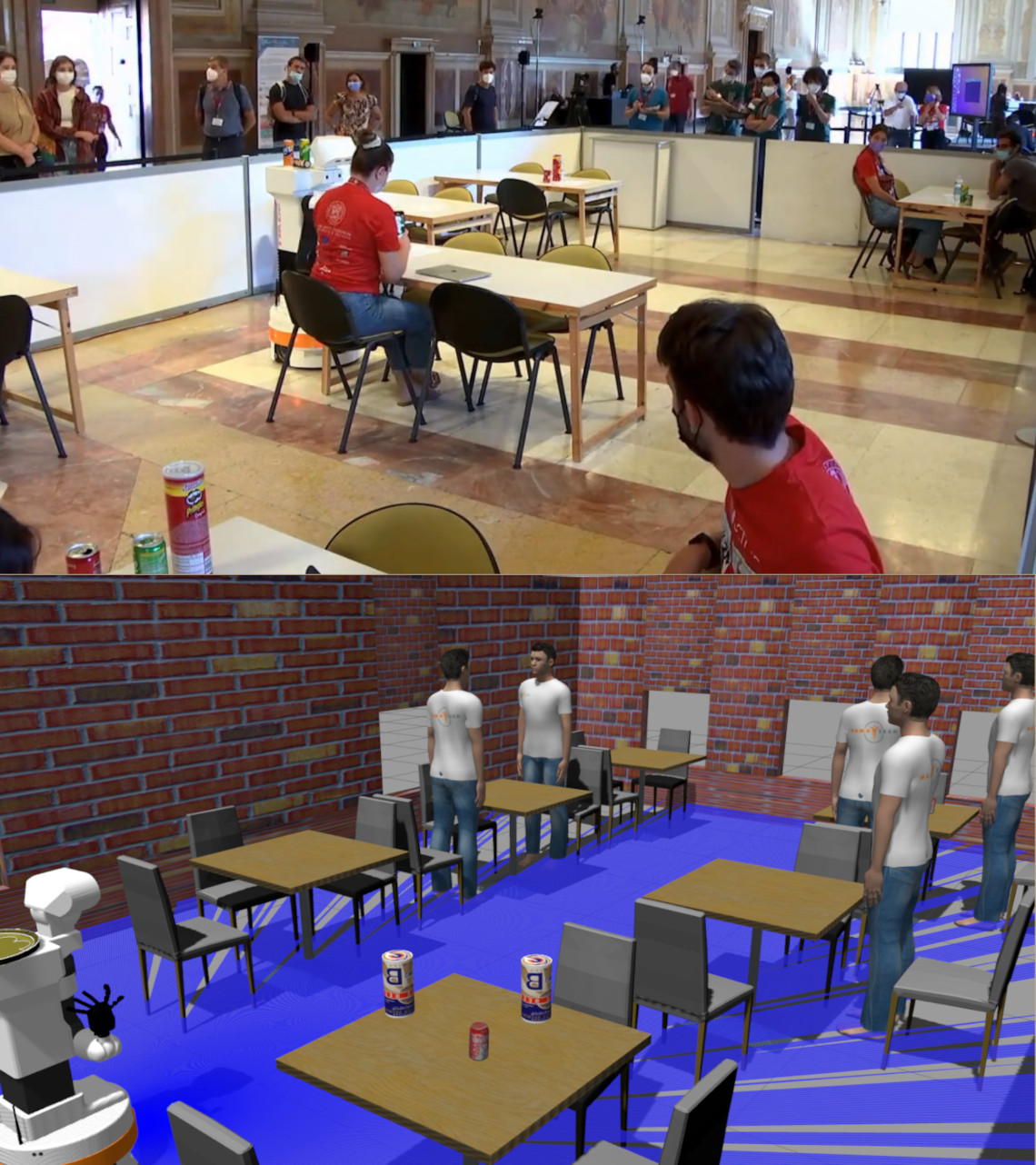}
    \caption{E01 \emph{Coffee Shop} in two of the proposed modalities: an interaction with the TIAGo robot during an on-site run in Bologna (upper image), and an execution in the corresponding simulated environment (lower image).}
    \label{fig:intro}
\end{figure}

SciRoc is a European project supporting the European Robotics League (ERL) whose aim is to bring ERL tournaments into smart cities and, up to now, this goal has been pursued through the organization of two SciRoc Challenges: in 2019 in Milton Keynes, UK, and in 2021 in Bologna, Italy. 
The objective of the Smart Cities Robotics Challenge (SciRoc) as a scientific competition is to provide \emph{Task Benchmarks (TBMs)} to teams, which allow for the measurement of performance, and to develop software and Human-Robot interaction (HRI) solutions. 
The second edition of Sciroc (SciRoc 2) 
has been structured in 5 \emph{episodes}\footnote{https://sciroc.org/2021-challenge-description/}, each consisting of a task to be performed through addressing specific research challenges.
In the Coffee Shop Episode (E01), in which the robot assists the staff of a coffee shop to take care of their customers, the robot is required to recognize and report the status of all tables inside the shop, to take orders from customers and to deliver objects to and from the customers’ tables.

The design of E01 episode of SciRoc 2, paid particular attention to HRI, verbal and non-verbal interaction, navigation, object detection and person detection as \emph{Functional Benchmarks (FBM)}. As compared with SciRoc 1, the new challenge of SciRoc 2 addressed the benchmarking of HRI functionality in simulated and real environments. In fact, 
due to the Covid-19 restrictions, E01 has been proposed in three settings: \textit{Physical, i.e., on-site with a real robot; Remote, i.e., remote with the robot on site; Simulated, i.e., (remote) simulated robot in a Gazebo environment}. Fig. \ref{fig:intro} shows a run in the on-site participation and a run in the simulated participation. 
Six teams participated in E01, using all the modalities offered.  This allowed teams to develop a single platform to test and develop the software for the three proposed settings. The multiple modalities of participation, represent a novelty in this kind of challenge especially considering that HRI is a required functionality: verbal and non-verbal interactions were part of each run of the Episode.
To this end, in all the modalities, special attention has been put in preserving \emph{real} interaction. 
We believe that our setup pushes forward the boundaries in having remote benchmark facilities to standardize the testing procedures of software. 

\begin{figure}
    \centering
    \includegraphics[width=0.99\columnwidth]{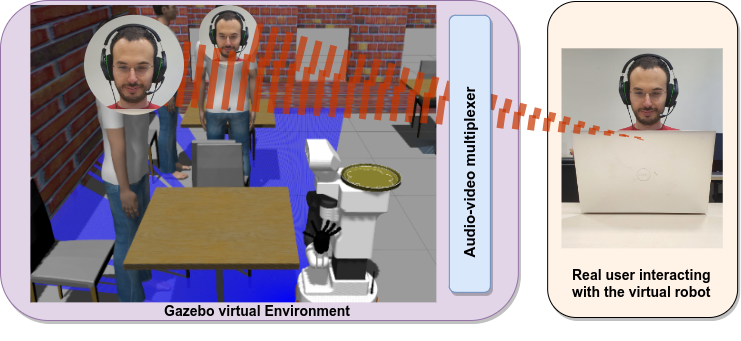}
    \caption{The simulated modality, executed with real persons, with the virtual robot able to switch between the simulated camera and the external webcam and microphone.}
    \label{fig:sim_with_real}
\end{figure}

\section{Related work}
\label{sec:related_work}
Competitions always played a significant role in the research in robotics, allowing to foster the development and deployment of new technologies and creating standard benchmark environments to evaluate the performances of the research approaches \cite{anderson_baltes_cheng_2011,yanco2015analysis,amigoni2015competitions}. In fact, as compared with a laboratory, the competition setting helps to improve the \emph{robustness} of the application and, hence, the reliability of the performance evaluated. 

The European Robotics League (ERL) approach to benchmarking experiments is based on the definition of two separate, but interconnected, benchmarking modalities: (1) \emph{Functionality Benchmarks (FBMs)}, that evaluate the performance of hardware/software modules dedicated to single, specific functionalities; and (2) \emph{Task benchmarks (TBMs)}, that evaluate the performance of integrated robotic systems executing tasks that need the interaction/composition of different functionalities \cite{fontana2017rockin}. The evaluation of the performances can be carried out at different levels, from the single robotic skills to the functionalities and, finally, to the overall TBMs. 
Among those, for the present work, we are interested in presenting the modalities that allowed to evaluate the Benchmarking during the SciRoc 2021 Competition. 


Before the Covid-19 pandemic, few robotic competitions have been held in \emph{remote participation} or \emph{simulated participation}. For instance, Robocup@Home simulation competition was held in 2013. After the Covid-19 pandemic, all robotic competitions have been transformed in \emph{remote} or \emph{simulated} modality. RoboCup Soccer Standard Platform League has been an example of this. In fact, two events have been organized within this league, during the pandemic: the GORE and the SPL RoboCup 2021 Worldwide. The German Open Replacement Event (GORE) 2021 has been organized as follows: all participants stayed at their premises (except for a few people supporting the organization) and were given remote access to a random subset of robots from the pool of a particular competition site for each game \cite{laue2021let}. The remote modality challenge was successful, sometimes, at the expense of maximum performance. A similar setup allowed the other competition, the RoboCup 2021 Standard Platform League (SPL), to play the matches in a distributed fashion with real robot games and remote participation. In fact, six fields with standard size were installed in different geographical locations. Participating teams sent their code to the local hosts, who were able to run the code on their own robots \cite{stone2021robocup}. 

Open Cloud Robot Table Organization Challenge (OCRTOC) 2020 competition provided a simulation environment to test the benchmarks for robotic grasping and manipulation, focusing on object rearrangement scenarios. In \cite{inamura2013development}, researchers present a SIGVerse simulation platform that enables social and embodied interaction with virtual agents. The virtual agent is asked to perform the clean-up task and the cooperative cooking with TBMs focusing on detection, recognition, manipulation, navigation and HRI. 

Our design of E01 episode of SciRoc 2, introduced \emph{HRI}, through verbal and non-verbal interaction, as an FBM. As compared with SciRoc 1, the new challenge of SciRoc 2 addressed the implementation of HRI in simulated and real environments, using the proposed architecture.

\section{Scenario and Procedure of E01}

The scenario of E01 of SciRoc 2 is shown in  Fig. \ref{fig:intro}. The robot assisted the staff of an environment resembling a real coffee shop to take care of their customers. 
Both the real and the simulated environments have been designed with a surface of around 70 square meters, containing 6 tables and a counter. The environments comprised people waiting to be served, customers already served, tables that need to be cleaned, and empty tables ready to receive new customers. The items that customers were able to order are among those typically sold by real coffee shops. Fig. \ref{fig:real} shows a picture of the robot in the real environment, ready to reach the customer to take the order while Fig. \ref{fig:simulated} shows the robot in the starting position of the simulated environment, at the beginning of a run.

Inside the environments of E01, the TIAGo\footnote{\url{https://pal-robotics.com/robots/tiago/}} robot from PAL Robotics has been chosen for both the real and the simulated version of E01. 
In the case of the real environment, a \textit{TIAGo} robot has been provided on site to support remote participation and on-site participation of teams without a robot. Remote participation has been allowed through the delivery of a Docker container to be deployed on site. 

In E01, the procedure is the same for all the modalities. The robot is placed at a starting location near the counter, then, the trial begins. The robot starts navigating around the shop and detects the status of all tables, reporting it to the counter. If a table is clean but with customers present at the table, the robot reaches it to accept an order from that table and, then, reports it to the kitchen. The interaction with customers occurs via spoken language. The robot, then, navigates to the counter to collect the items and deliver them to that table. The robot does not require manipulation. Hence, the items are taken by the customers from the robot tray. On each run, the order provided by the counter has one of the items missing or incorrect and the robot must identify and correct the mistake. A run terminates when the robot has delivered the order to the table. A time limit ends the episode if the robot gets stuck.




\begin{figure}
    \centering
    \includegraphics[width=0.99\columnwidth]{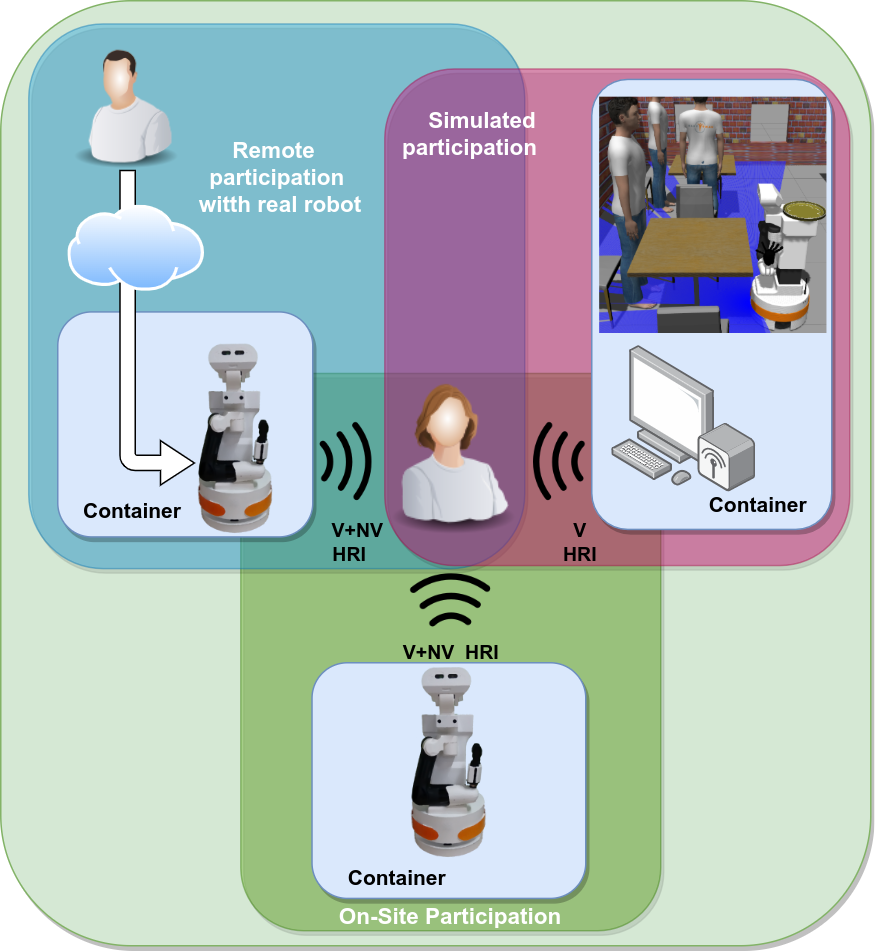}
    \caption{A schema of the three modalities highlighting the HRI capabilities. The container provided to the participants of E01 is the same for all the modes. In the three cases, the HRI has been executed with real persons, guaranteeing \textit{verbal (V)} and \textit{non-verbal (NV)} interactions.}
    \label{fig:modalities}
\end{figure}

\section{SciRoc E01 as a Multiplatform Benchmark System}
\label{sec:cooperation}

The SciRoc 2 competition has been organized and hosted during the Covid-19 pandemic. The effort made to cope with the restrictions caused by the pandemic led to novel technical solutions that are described in this section. We have prepared E01 code base to allow teams to use a single Docker container capable of hosting and run the software in the three modalities proposed for the participation in the challenge, as represented in Fig. \ref{fig:modalities}. This container, developed with PAL Robotics, has been provided with \textit{ROS Melodic}, the set of TIAGo drivers, a complete navigation stack, a text-to-speech synthesizer, a simulated TIAGo and a virtual environment developed within the Gazebo simulator. The use of the TIAGo robot helped in standardizing the code base since its software is entirely based on ROS. This allowed us to implement a simulated testing environment and to simplify the deployment of participating teams' software that can run on the robot and on the laptops connected to the robot.  The whole image is publicly released at the following link: \url{https://gitlab.com/competitions4/sciroc/dockers/-/tree/master}.

\begin{figure}
    \centering
    \includegraphics[width=1.0\columnwidth]{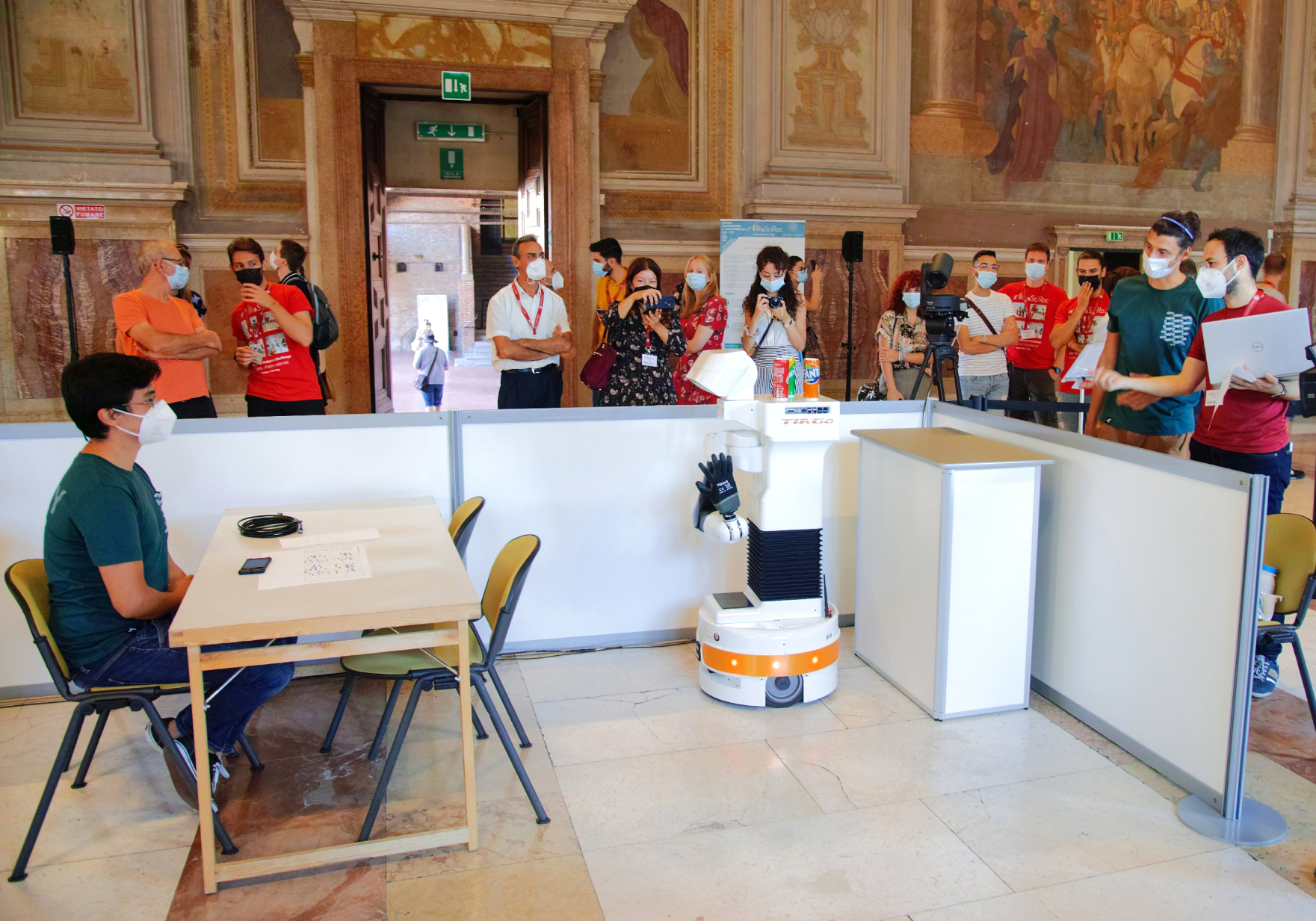}
    \caption{E01, on-site with the TIAGo robot, the counter and a customer in the coffee shop at the beginning of a run, in Bologna.}
    \label{fig:real}
\end{figure}

\paragraph{SciRoc E01 - On-site Participation}
The on-site procedure required the teams to be in person in Bologna at the competition site, with their own robot or the shared TIAGo robot provided by the organizers. 
The map of the environment has been provided and the images of the coffee shop items were delivered to teams some weeks before the competition. 
In order to simplify the deployment of the software on the robot, each team could extend the provided container by adding at least: \textit{object recognition}, \textit{person detection}, \textit{dialog management} and \textit{robot behaviors}. The deployment procedure has been carried on by loading the Docker image on the robot before each run. A picture of a run is shown in Fig. \ref{fig:real}.
In this case, verbal and non-verbal interactions occurred totally through the onboard sensors of the robot. 
A timelapse of a run of the episode can be seen at the following link: \url{https://youtu.be/UckQEKtYep8}. 

\paragraph{SciRoc E01 - Remote Participation in the Real Scenario}
For remote participation, a laptop has been prepared to host the containers of the deploying from remote. 
For the remote participants, beyond the map, the organizers had to provide also a set of bags of the items to be recognized and a \textit{Rosbag} taken with the robot navigating in the environment. 
Even in this case, verbal and non-verbal interactions occurred only with the hardware of the robot.

\begin{figure}
    \centering
    \includegraphics[width=1.0\columnwidth]{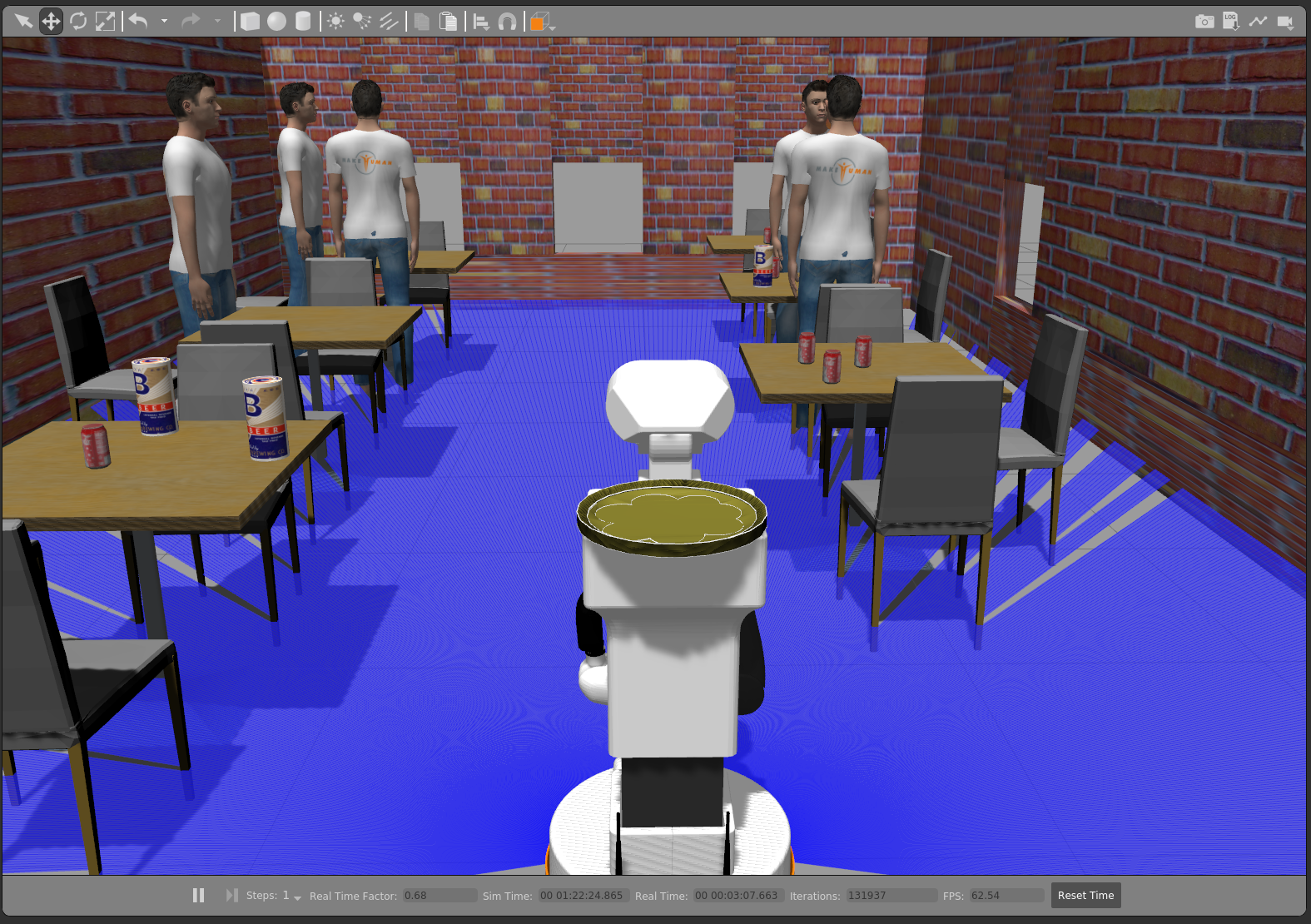}
    \caption{E01 Simulated. The figure represents one of the starting configurations of the simulated episode, running in the Gazebo simulator.}
    \label{fig:simulated}
\end{figure}

\begin{figure}
    \centering
    \includegraphics[width=1.0\columnwidth]{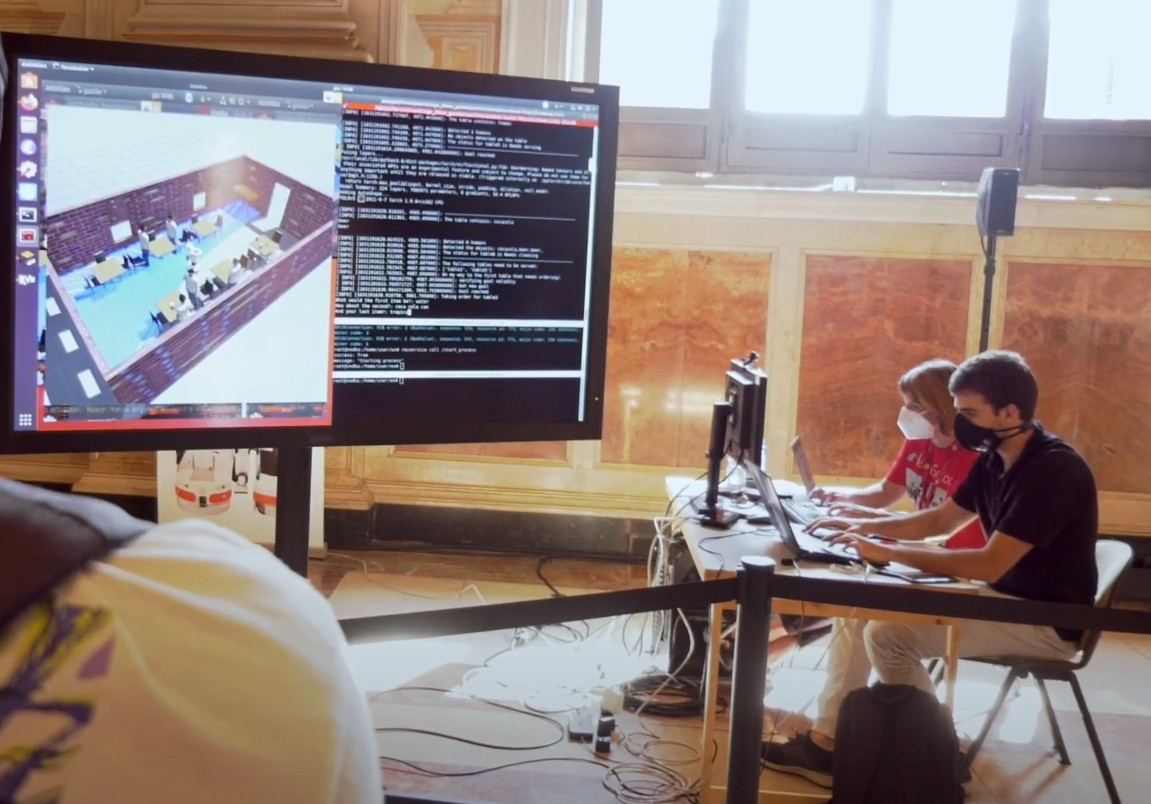}
    \caption{E01 in a remote simulated run with real customers, placed in front of the small monitor, interacting with the simulated robot shown in the external screen.}
    \label{fig:remote_run}
\end{figure}

\paragraph{SciRoc E01 - Simulation} 
The software of the remote simulated participation has been provided with two graphical tools: a Gazebo environment that resembles the real one, as shown in Fig. \ref{fig:simulated}, and an RViz custom interface developed by Pal Robotics.
In this environment, the simulated TIAGo has the same capabilities of the real one and some provided ROS services allow to perform automatically the actions over the items that, in the real world, are performed by the bartender and the customers. 
An important feature of this code base is the possibility to preserve the HRI procedure: by taking advantage of the camera and the microphone of the computer that hosts the simulation, we blended the simulation with the real environment and, the robot in the simulation could interact with the real customers at the competition site. In fact, during the runs of the simulation, the robot was able to use the simulated camera to perform navigation and object detection tasks, but it also used the webcam and the microphone of the host computer to interact with customers, as depicted in Fig. \ref{fig:sim_with_real}. Fig. \ref{fig:remote_run} shows a run of the remote simulated modality at the competition site: the external screen shows the simulated scenario while real customers that can interact with the simulated robot are placed in front of the webcam and microphone of the computer, guaranteeing verbal interactions with real customers. 

\section{Evaluation}
In E01, HRI was one the TBMs functionalities.
The evaluation of TBMs of a robot is based on the concept of \textit{performance classes} employed in the ERL competitions. The performance class of a robot is determined by the number of achievements that the robot collects during its execution of the assigned task. 
In all the proposed modalities, verbal interaction with customers has been an essential achievement in order to complete the challenge. In the cases of the \emph{on-site and remote participation with a real robot}, even the non-verbal interaction (for instance, proxemics between robot and humans
) has been part of the evaluated functionalities.
In the simulated setting, the structure of the virtual environment did not allow the disposal of non-verbal interaction. Nevertheless, verbal interaction has been the most challenging functionality. 
The choice to have a real HRI procedure between the simulated robot and real persons facing the computer, has given rise to have a declared issues with the models of the microphone and the camera. In fact, teams expressed their need to have a standardized hardware for the HRI procedure.

\section{Conclusions
}
\label{sec:conclusion}

 
The Episode 1 of the SciRoc 2 competition allowed us 
to introduce new modalities to standardize benchmarks for robotic applications in Human-Robot Interaction scenarios. E01 competition provided to participants a standard benchmarking facility, aiming at reducing the gap between the simulation scenario and the real-world scenario in terms of HRI capabilities. In fact, it allowed the deployment of the developed software on different modalities: remotely simulated, and remotely and in-person on a real benchmarking environment created on purpose at the competition site. In all of them, we preserved real human-robot interactions. For the remote simulated setup, we proposed and adopted an architecture that blends the simulated perceptions and the perceptions coming from the real environment. Despite the worldwide restrictions imposed by the Covid-19 pandemic, the developed architecture allowed the hosting of a competition among six teams that took advantage of all the proposed modalities. The competition has been fully decentralized with teams participating remotely by sending their software in a container and making it interact with the customers in presence.
This pushes forward the boundaries in having remote benchmark facilities to standardize the testing procedures of software, blending real and simulated environments with a shared set of metrics.




\begin{acks}
We wish to thank PAL Robotics 
for the technical support in the development of the software architecture used 
in this competition.
\end{acks}

\bibliographystyle{ACM-Reference-Format}
\bibliography{main}

\end{document}